\documentclass[sigconf]{acmart}

\def\BibTeX{{\rm B\kern-.05em{\sc i\kern-.025em b}\kern-.08emT\kern-.1667em\lower.7ex\hbox{E}\kern-.125emX}}
\copyrightyear{2018}
\acmYear{2018}
\setcopyright{acmlicensed}
\acmConference[Woodstock '18]{Woodstock '18: ACM Symposium on Neural Gaze Detection}{June 03--05, 2018}{Woodstock, NY}
\acmPrice{15.00}
\acmDOI{10.1145/1122445.1122456}
\acmISBN{978-1-4503-9999-9/18/06}

\usepackage{times}
\usepackage{soul}
\usepackage{url}
\usepackage{hyperref}
\usepackage{inputenc}
\usepackage{caption}
\usepackage{graphicx}
\usepackage{amsmath}
\usepackage{mathtools}
\usepackage{booktabs}
\usepackage{algorithm}
\usepackage{algorithmic}
\usepackage{paralist}
\usepackage{xcolor}
\begin{document}

\title{Fake Node Attacks on Graph Convolutional Networks}

\author{Xiaoyun Wang}
\email{xiywang@ucdavis.edu}
\affiliation{%
  \institution{University of California, Davis}
}

\author{Minhao Cheng}
\email{ mhcheng@g.ucla.edu}
\affiliation{%
  \institution{University of California, Los Angeles}
}
\author{Joe Eaton}
\email{featon@nvidia.com}
\affiliation{%
  \institution{NVIDIA}
}
\author{Cho-Jui Hsieh}
\email{chohsieh@cs.ucla.edu}
\affiliation{%
  \institution{University of California, Los Angeles}
}
\author{S. Felix Wu}
\email{wu@cs.ucdavis.edu}
\affiliation{%
  \institution{University of California, Davis}
}

\begin{abstract} 
In this paper, we study the robustness of graph convolutional networks (GCNs). 
Previous work have shown that GCNs are vulnerable to adversarial perturbation on adjacency or feature matrices of existing nodes; however, such attacks are usually unrealistic in real applications. 
For instance, in social network applications, the attacker will need to hack into either the client or server to change existing links or features. 
In this paper, we propose a new type of ``fake node attacks'' to attack GCNs by adding malicious fake nodes. This is much more realistic than previous attacks; in social network applications, the attacker only needs to register a set of fake accounts and link to existing ones. 
To conduct fake node attacks, 
a greedy algorithm is proposed to generate edges of malicious nodes and their corresponding features
aiming to minimize the classification accuracy on the target nodes. 
In addition, we introduce a discriminator to classify malicious nodes from real nodes, and propose a Greedy-GAN attack to simultaneously update the discriminator and the attacker, to make malicious nodes indistinguishable from the real ones.  
Our non-targeted attack decreases the accuracy of GCN down to 0.03, and our targeted attack reaches a success rate of 78\% on a group of 100 nodes, and 90\% on average for attacking a single target node.

\end{abstract} 


\maketitle

\definecolor{rd} {rgb} {1.0,0.0,0.0}
\definecolor{lg} {rgb} {0.7,0.3,0.9}
\definecolor{db} {rgb} {0.0,0.0,0.7}
\definecolor{dg} {rgb} {0.0,0.7,0.0}
\definecolor{dr} {rgb} {0.7,0.0,0.0}
\definecolor{gr} {rgb} {0.8,0.4,0.1}

\newcommand{\TODO } [1] {{{\color{rd}[TODO] #1}}}
\newcommand{\XY } [1] {{{\color{db}[XY] #1}}}
\newcommand{\MO   } [1] {{{\color{lg}[MO]#1}}}
\newcommand{\CH   } [1] {{{\color{dg}[CH]#1}}}
\newcommand{\YC   } [1] {{{\color{db}[YC]#1}}}
\newcommand{\MH   } [1] {{{\color{gr}[MH]#1}}}

\newcommand{\final} {1}

\ifthenelse{\equal{\final}{1}} {\renewcommand{\XY }[1]{}}{}
\ifthenelse{\equal{\final}{1}} {\renewcommand{\MO }[1]{}}{}
\ifthenelse{\equal{\final}{1}} {\renewcommand{\CH }[1]{}}{}
\ifthenelse{\equal{\final}{1}} {\renewcommand{\YC}[1]{}}{}
\ifthenelse{\equal{\final}{1}} {\renewcommand{\MH}[1]{}}{}

\begin{abstract} 
In this paper, we study the robustness of graph convolutional networks (GCNs). 
Previous work have shown that GCNs are vulnerable to adversarial perturbation on adjacency or feature matrices of existing nodes; however, such attacks are usually unrealistic in real applications. 
For instance, in social network applications, the attacker will need to hack into either the client or server to change existing links or features. 
In this paper, we propose a new type of ``fake node attacks'' to attack GCNs by adding malicious fake nodes. This is much more realistic than previous attacks; in social network applications, the attacker only needs to register a set of fake accounts and link to existing ones. 
To conduct fake node attacks, 
a greedy algorithm is proposed to generate edges of malicious nodes and their corresponding features
aiming to minimize the classification accuracy on the target nodes. 
In addition, we introduce a discriminator to classify malicious nodes from real nodes, and propose a Greedy-GAN attack to simultaneously update the discriminator and the attacker, to make malicious nodes indistinguishable from the real ones.  
Our non-targeted attack decreases the accuracy of GCN down to 0.03, and our targeted attack reaches a success rate of 78\% on a group of 100 nodes, and 90\% on average for attacking a single target node.

\end{abstract} 
\section{Introduction}
\label{introduction}

Graph data has been widely used in many real world applications,  such as  social networks (Facebook and Twitter), biological networks (protein or gene interactions), as well as  attribute graphs (PubMed and Arxiv) \cite{Grover:2016:NSF,pinsage,breastcancer}. 
Node classification is one of the most important tasks on graphs---given a graph with labels associated with a subset of nodes, predict the labels for the rest of the nodes. 
For this node classification task, deep learning models on graphs, such as Graph Convoltional Networks (GCNs), have achieved state-of-the-art performance  \cite{DBLP:journals/corr/KipfW16}. 
Moreover, GCNs have wide applications in cyber security, where they can learn a close-to-correct node labeling semi-autonomously. This reduces the load on security experts and helps to manage networks that add or remove nodes dynamically, such as, WiFi networks in universities and web services in companies. 

The power of GCN relies  on the graph convolution operation, which constructs each node's embedding based on its neighbors' embeddings. However this also leads to the  concern on the robustness---Is it possible to perturb the links and embeddings of a small subset of nodes to affect the GCN's behavior on other nodes? 
To answer this question, 
\cite{netattack,pmlr-v80-dai18b}
develop algorithms to attack GCNs, showing that by altering a small amount of edges and features, the classification accuracy of a  GCN can be reduced to chance level. However, changing edges or features associated with existing nodes is usually impractical.
For example, in social network applications, an attacker has to login to the users' accounts to change existing connections and features, and gaining login accesses is almost impossible. 
In comparison, adding fake nodes that correspond to fake accounts or users, can be much easier in practice. 
But the key question is that  {\it can we interfere the classification results of existing nodes by adding fake nodes to the network? }
We answer this question affirmative by introducing novel algorithms to design fake nodes that successfully reduce GCN's performance on existing nodes.

To design the adjacency and feature matrices associated with fake nodes, we have to address two challenges. First, the edges and features are usually discrete 0/1 variables. 
Although there have been many algorithms proposed for attacking image classifiers, such as FGSM, C\&W and PGD attacks \cite{fgsm,cw,pgd}, they all assume continuous input space and cannot be directly applied to problems with discrete input space. 
Second, it is not easy to make the fake nodes ``look'' like the real ones. There are two aspects for realness of fake nodes, i) in graph structure  and ii) in features. 
For example, if we add a fake node that connects to all existing nodes, 
the system can easily detect and disable such fake node; and if a fake node features are all 1s or very large or small value in continuous cases, the system can also easily detect it. 
In this paper, we propose two algorithms, Greedy attack and Greedy-GAN attack, to address these two challenges. Our contributions can be summarized below: 
\begin{itemize}
\item To the best of our knowledge, this is the first paper studying how to add fake nodes to attack GCNs. We do not need to manipulate existing nodes' adjacency and feature matrices. 
\item We propose a Greedy attack algorithm to address the discrete input space problem in designing fake nodes' adjacency and feature matrices. 
\item We introduce a discriminator to classify fake nodes from real nodes, and propose a Greedy-GAN algorithm to simultaneous optimize the discriminator and the attacker. Despite a slight lower successful rate, this approach can make fake nodes harder to detect. 
\item We conduct experiments on several real datasets. 
For non-targeted attack, we are able to degrade the classifier's accuracy to less than $10\%$ on both Cora and Citeseer datasets, and for targeted attack we can achieve a $>70\%$ success rate when attacking either a single node or a group of nodes. 

\CH{summarize results}
\end{itemize}

\section{Related Work}
\label{related}


\paragraph{Adversarial Attacks.}
Adversarial examples for computer vision have been studied extensively. ~\cite{fgsm} discovered that deep neural networks are vulnerable to adversarial attacks---a carefully designed small perturbation can easily fool a neural network. 
Several algorithms have been proposed to generate adversarial examples for image classification tasks, including FGSM \cite{fgsm}, IFGSM ~\cite{ifgsm}, C\&W attack  ~\cite{cw} and PGD attack ~\cite{pgd}. 
In the black-box setting, it has also been reported that an attack algorithm can have high success rate using finite difference techniques  \cite{zoo,cheng2018query}, and several algorithms are recently proposed to reduce query numbers  \cite{limitequrry,querylimit2}. 
Transfer attack \cite{DBLP:journals/corr/PapernotMG16} and ensemble attack  \cite{DBLP:journals/corr/TramerKPBM17} can also be applied to black-box setting, with lower successful rate but less number of queries.
\CH{Can remove the following sentences. } \XY{understand} 
Besides attacking image classification, CNN related attacks have also been explored. A typical usage is attacking semantic segmentation and object detection \cite{DBLP:conf/iccv/MetzenKBF17_i21,cheng2018query,DBLP:journals/corr/abs-1711-09856_i22,DBLP:conf/iccv/XieWZZXY17_i23,DBLP:journals/corr/abs-1712-02494_i24,DBLP:journals/corr/abs-1712-08062_i25}. Image captioning \cite{DBLP:journals/corr/abs-1712-0205_i26} and visual QA \cite{DBLP:journals/corr/abs-1709-08693_i27} could also be attacked. 

Most of the above-mentioned work are focusing on problems with continuous input space (such as images). 
When the input space is discrete, attacks become discrete optimization problems and are much harder to solve. 
This happens in most natural language processing (NLP) applications. 
%
For text classification problem, Fast Gradient Sign Method (FGSM) is firstly applied by \cite{DBLP:conf/milcom/PapernotMSH16(33)}.  Deleting important words \cite{DBLP:journals/corr/LiMJ16a_(34)}, replacing or inserting words with typos and synonyms \cite{DBLP:journals/corr/SamantaM17(35),DBLP:journals/corr/LiangLSBLS17(36)}. For black box setting, \cite{DBLP:conf/sp/GaoLSQ18_i37} develops score functions to find out the words to modify. \cite{DBLP:conf/emnlp/JiaL17_i38} adds misleading sentences to fool the reading comprehension system.  \cite{DBLP:journals/corr/abs-1710-11342_i39} uses GAN to generate natural adversarial examples. \cite{whitebox} and \cite{seq2sick} attacks machine translation system Seq2Seq by changing words in text. 

\paragraph{Graph Convolutional Neural Networks (GCNs).}
GCN has been widely used for node classification and other graph-based applications. 
The main idea of GCN is to aggregate associated information from a node and its neighbors using some aggregation functions. They  train the aggregation steps and the final prediction layer end-to-end to achieve better performance than traditional approaches. 
There are several variations of GCNs proposed recently \cite{DBLP:journals/corr/KipfW16,cln,DBLP:conf/nips/DefferrardBV16,graphsage,fastgcn,pinsage}. 

The wide applicability of GCNs motivates recent studies about their robustness. 
\cite{netattack,pmlr-v80-dai18b} recently proposed algorithms to attack GCNs by changing existing nodes' links and features. 
\cite{netattack} developed an FGSM-based method that optimizes a surrogate model to choose the edges and features that should be manipulated. 
\cite{pmlr-v80-dai18b} 
proposed several attacking approaches including, gradient ascent, Genetic algorithm and reinforcement learning. 
\XY{should we use algorithm names in their paper? like GradArgmax ?}
They employed a gradient ascent method to change the graph structure in the white-box setting.
\CH{Not clear, are they using RL or using gradient ascent?}
\XY{Song use several different methods to attack, the gradient is white box setting, and it is the one we compared in experiments}
However, their settings are different from ours in the following aspects: 
1) Our methods have different constraint than the previous works. They considered altering edges or features of existing nodes while we develop  algorithms to add fake nodes to interfere the performance of GCN. In their scenarios keeping graph structure perturbations unnoticeable is important. While in our adding fake nodes scenario, only keeping structure perturbations unnoticeable is far from enough. Keeping fake nodes features realness is very important, if a fake node features are very unreal, the system can also easily detect it and disable this fake node. 
\XY{song's paper mentioned changing to target is not hard, but did not do experiments on it.}
2) They attacked nodes one-by-one while our algorithms can find a small perturbation to attack a group of nodes.



\section{Preliminary}
\label{prelim}
We consider the node classification problem using GCN. 
Given an adjacency matrix $A\in R^{n\times n}$,  feature matrix $X\in R^{n\times d}$, and a subset of labeled nodes, the goal is to predict the labels of all the nodes in the graph. We use $n$ to denote number of nodes, $P$ to denote number of labels, 
and $d$ to denote the dimension of features (each row of $X$ corresponds to features of a node). 
There are several variations of GCNs, but we consider one of the most common approaches introduced in  \cite{DBLP:journals/corr/KipfW16}. 
Starting from $H^{0}=X$, GCN  uses the following rule to iteratively aggregate features from neighbors: 
\begin{equation}
H^{(k+1)} = \sigma ( \hat{A}  H ^{(k)} W^{(k)}),
\end{equation}
where $\hat{A}$ is the normalized adjacency matrix and $\sigma$ is the activation function.  
There are two common ways for normalizing $A$. The original GCN paper applied {\bf symmetric normalization}, which sets
$\tilde{A}=A+I$ (adding 1 to all the diagonal entries) and then normalize it by 
$\hat{A}=\tilde{D}^{-\frac{1}{2}} \tilde{A} \tilde{D}^{-\frac{1}{2}}$ where $\tilde{D}$ is a diagonal matrix with $\tilde{D}_{i,i} = \sum_{j}^{} {\tilde{A}_{ij}}$. Another choice for $\hat{A}$ is to normalize the adjacency matrix by rows, leading to $\hat{A}=\tilde{D}^{-1} \tilde{A}$, which is referred to as {\bf row-wise normalization}.  \XY{changed to correct -1}

We set  $\sigma(x) = \text{ReLU} (x) = \max (0, x)$, which is the most common choice in practice.
For a GCN with $K$ layers, after getting the top-layer feature $H^{(K)}$, a fully connected layer with soft-max loss is used for classification. 
Note that there are two types of normalization have been used in GCN.

A commonly used application is to apply two-layer GCN to semi-supervised node classification on graphs~\shortcite{DBLP:journals/corr/KipfW16}. The model could be simplified as:
\begin{align}
Z =& \text{softmax} (\hat A  \sigma (\hat A X W^{(0)}) W^{(1)}) \nonumber \\
:= & \text{softmax}(f(X, A)). 
\label{eq:two_layer}
\end{align}
Note that 
each row of 
$f(X, A)\in R^{n\times P}$ is the logit output for a node over $P$ labels, and the softmax is taken on each row of the matrix. 

For simplicity, we will assume our target network is structured as~\eqref{eq:two_layer}, but in general our algorithm can be used to attack GCNs with more layers. 

\section{Attack Algorithms}
\label{attack}
In this section, we will describe our ``fake node'' attacks. We will describe both a non-targeted attack,  which aims to lower the accuracy of all  existing nodes uniformly, and a targeted attack, which attempts to force the GCN to give a desired label to nodes.
Instead of manipulating the feature and adjacency matrices of existing nodes, we insert $m$ fake nodes with corresponding fake features into the graph. After that, the adjacency matrix is
$
    A' = \begin{bmatrix}
A & B^T \\ 
B & C 
\end{bmatrix}
$
and the feature matrix becomes :
$
    X' = \begin{bmatrix}
X \\ 
X_{fake} 
\end{bmatrix}
$. Note that $A$ is the original adjacency matrix and $X$ is the original feature matrix. 
Starting from $B=0, C=I$, our goal is to design $B, C, X_{fake}$ to achieve the desired objective (e.g., lower the classification accuracy on existing nodes). 


\subsection{Non-targeted Attack}\CH{The first char needs to be capital} \XY{changed}

The goal of non-targeted attack is to lower the classification accuracy on all the existing nodes by designing features and links of fake nodes. 
We use the accuracy of GCN to measure the effectiveness of attacks. We will present two different algorithms to attack GCNs: Greedy attack that updates links and features one by one, and Greedy-GAN attack that uses a discriminator to generate unnoticeable features of fake nodes. 



\subsubsection{Greedy Attack }
We define the  objective function of the attack as
\begin{equation}
    \!\!J(A' , X') \!=\!
    \!\sum_{i\in S} \!
   \left( \max\big([f(X', A')]_{i,:}\big) \!-\! 
[f(X', A')]_{i, y_i}\!  \right)\!, 
\end{equation}
where $y_i$ is the correct label of node $i$; $S$ is a group of $m$ target nodes  $S=\{v_1, \dots, v_{m}\}$. In this objective function, if the largest logit of node  $i$ is not the correct label $y_i$, it will encounter a positive score; otherwise the score will be zero. 
We then try to solve the following optimization problem to form the fake nodes: 
\begin{equation}
\arg\max_{B, C, X_{fake}}
J(A', X')  \ \ \ 
    \text{ s.t. } \|B\|_0 + \|C\|_0 + \|X_{fake}\|_0 \leq T, 
    \label{eq:greedy}
\end{equation}
where $\|\cdot\|_0$ denotes number of nonzero elements in the matrix. Also, we assume $B, C, X_{fake}$ can only be 0/1 matrices. 
Unlike images, graphs have discrete values in the adjacency matrix, and in many applications the feature matrix comes from indicator of different categories. 
For some attribute graphs feature matrices are generated using word count; in practice, feature matrices are from categorical features, thus they are often discrete. 

Therefore, gradient-based techniques such as FGSM and PGD cannot be directly applied. 
Instead, we propose a greedy approach---starting from $B, X_{fake}$ all being zeros, and $C$ being identity matrix $I$, we add one feature and one edge at each step. To add a feature, we find the the maximal element in $ \nabla_{X_{fake}} J (A',X')$ and  turn it into nonzero. Similarly, we find the maximal element in $ \nabla_{B, C} J (A',X')  $ and add the entry to the adjacency matrix.
The Greedy attack is presented in Algorithm 1. 
\begin{algorithm}[tb]
   \caption{Greedy Attack}
   \label{alg:fgsm}
\begin{algorithmic}
    \STATE {\bfseries Input:} Adjacency matrix $A$; feature matrix $X$; A classifier $f$ with loss function $J$; number of iterations $T$. 
   \STATE {\bfseries Output:} Modified graph and features $G' = (A',X')$ after adding fake nodes. 
   \STATE {\bfseries for:} t = 0 to $T-1$  \bfseries do
   \STATE \hspace{\algorithmicindent} Let $e^* =(u^*,v^*) \leftarrow \arg \max  \nabla_{B,C} J (A',X') $ \\
   \STATE\hspace{\algorithmicindent} \hspace{\algorithmicindent} $G_{B, C}^{(t+1)} \leftarrow  G_{B,C} ^{(t)} + e^* $ 
   \STATE \hspace{\algorithmicindent} Let $f^* =(u^*,i^*) \leftarrow \arg \max  \nabla_{X_{fake}} J (A',X') $ \\
      \STATE\hspace{\algorithmicindent} \hspace{\algorithmicindent} $G_{X_{fake}}^{(t+1)} \leftarrow  G_{X_{fake}}^{(t)} + f^* $ 

   \STATE return: $G^{(t)} $
   
\end{algorithmic}
\end{algorithm}
In the algorithm, when adding links and features, we make sure that there is no such a link or feature before adding. Also, although in Algorithm~\ref{alg:fgsm} we conduct one feature update and one edge update at each iteration, in practice we can adjust the frequency according to the data distribution. 
For example, if the original adjacency matrix has twice nonzero elements than the feature matrix, we can update two elements in the adjacency matrix and one element in the feature matrix at every iteration.

\subsubsection{Greedy-GAN Attack}


Unlike changing on existing graphs methods \cite{pmlr-v80-dai18b} \cite{netattack}, adding fake nodes on graphs is faced with different "unnoticeable" definition, and it has different constraint from the existing works on nodes features. 
Making edges perturbation unnoticeable is needed but it is far from enough for adding fake nodes setting. For example, if add nodes' features are very different from the existing nodes. It will be detected by data cleaning easily. Facing such a different constraint, ensuring fake nodes' features realistic becomes crucial. 
Next, we will present the attack based on the idea of Generative Adversarial Network (GAN).

The main idea is to add a discriminator to generate fake features that are similar to the original ones. 
To achieve this goal, we first design a neural network with two fully connected layers plus a softmax layer as the discriminator, which can be written as
\begin{equation}
    D(X') =  \text{softmax} (\sigma(X'W^{(0)})W^{(1)}),  
\end{equation}
where softmax works on each row of the output matrix. 
Each element in $D(X')$ indicates whether the discriminator classifies the node as real or fake. 

We aim to generate fake nodes with features similar to the  real ones to fool the discriminator. 
Since the output of discriminator is binary, we use binary cross entropy loss defined by $L(p, y) = -(y \log (p) + (1-y) \log (1-p) )$, where $y$ is binary indicator of the ground-truth (real or fake nodes), 
and $p$ is the predicted probability by our discriminator.
Then we solve the following optimization problem :
\begin{equation}
    \arg\max_{B,C, X_{fake}} \min_{D}(\underbrace{J (A',X') + c* L(D(X'), Y)}_{Q}),
    \label{eq:greedy_gan}
\end{equation} \XY{we want to maximize the loss J(), and maximize the loss of D() }
\CH{need to say how we define $L$; $\max_{?, ?}$ (do we also update the parameter of $D$?} \XY{added}
where $Y$ is the ground-truth (real/fake) indicator for nodes and 
$c$ is the parameter determine with the weight of discriminator and the GCN performance. For example, if $c$ is large, the objective function is dominated by the discriminator, so the node features generated will be very close to real ones but with a lower attack successful rate. We will discuss more in Section \ref{diff_c_subsect}. 

We adopt the GAN-like framework to train both features/adjacency matrices and discriminator parameters iteratively. In experiments, 
at each epoch we conduct $100$ greedy updates for $B, C, X_{fake}$ and then $10$ iterations of $D$ updates. 
The Greedy-GAN algorithm is presented in Algorithm \ref{alg_cw}. 
Greedy-GAN supports both adding and dropping links and features. In the algorithm, we add or drop elements according to the absolute gradient of elements, and the one with larger absolute value will be chosen. 

The time complexity of Greedy and Greedy-GAN per iteration for updating edges of fake nodes costs $O(|V|*k +k^2)$, where $k$ is the number of added fake nodes; updating features of fake nodes costs $O(k*N_{features})$, $N_{features}$ is number of features each node.  

\begin{algorithm}[tb]
   \caption{Greedy-GAN Attack \CH{Modify the algorithm structure (discuss)}}
   \label{alg_cw}
\begin {algorithmic}
   \STATE {\bfseries Input:} Adjacency matrix $A$; feature matrix $X$; A classifier $f$ with loss function $J$; Discriminator $D$ with loss function $L$; number of outer iterations $T_{outer}$, and inner iterations $T_{inner}$. 
   \STATE {\bfseries Output:} Modified graph $G' = (A',X')$ after adding fake nodes. 
   \STATE {\bfseries for:} $t_1 = 0$ to $T_{outer}-1$  {\bfseries do}
   \STATE \hspace{\algorithmicindent} {\bfseries for:} $t_2 =0$ to $T_{inner} -1$ {\bfseries do}
   \STATE \hspace{\algorithmicindent}\hspace{\algorithmicindent} {\bfseries Let} $e^*_{add} =(u^*_{add},v^*_{add}) \leftarrow \arg \max \nabla_{B,C}  Q$ \\
    \STATE \hspace{\algorithmicindent}\hspace{\algorithmicindent}\hspace{\algorithmicindent}  $e^* _{drop } = (u^*_{drop},v^*_{drop}) \leftarrow \arg \min  \nabla_{B,C} Q $
   \STATE\hspace{\algorithmicindent}\hspace{\algorithmicindent} {\bfseries if} $|\nabla_{B,C} Q_{e^*_{add}}| > |\nabla_{B,C} Q_{e^*_{drop}}|  $ :
      \STATE\hspace{\algorithmicindent} \hspace{\algorithmicindent}\hspace{\algorithmicindent} $G_{B, C}^{} \leftarrow  G_{B, C}^{} + e^*_{add} $ 
   \STATE \hspace{\algorithmicindent}\hspace{\algorithmicindent} {\bfseries else}:
      \STATE \hspace{\algorithmicindent}\hspace{\algorithmicindent}\hspace{\algorithmicindent} $G_{B, C}^{} \leftarrow  G_{B, C}^{} - e^*_{drop} $ 
      \STATE \hspace{\algorithmicindent}\hspace{\algorithmicindent}{\bfseries Let} $f^*_{add} =(u^*_{add},i^*_{add}) \leftarrow \arg \max  \nabla_{X_{fake}} Q $ \\
    \STATE \hspace{\algorithmicindent}\hspace{\algorithmicindent}\hspace{\algorithmicindent}  $f^* _{drop } = (u^*_{drop},i^*_{drop}) \leftarrow \arg \min  \nabla_{X_{fake}} Q $
      
   \STATE \hspace{\algorithmicindent}\hspace{\algorithmicindent} {\bfseries if} $|\nabla_{X_{fake}} Q_{f^*_{add}}| > |\nabla_{X_{fake}} Q_{f^*_{drop}}|  $ :
      \STATE\hspace{\algorithmicindent}\hspace{\algorithmicindent} \hspace{\algorithmicindent} $G_{X_{fake}}^{} \leftarrow  G_{X_{fake}}^{} + f^*_{add} $ 
   \STATE \hspace{\algorithmicindent}\hspace{\algorithmicindent} {\bfseries else}:
      \STATE\hspace{\algorithmicindent} \hspace{\algorithmicindent}\hspace{\algorithmicindent} $G_{X_{fake}}^{} \leftarrow  G_{X_{fake}}^{} - f^*_{drop} $ 
    \STATE\hspace{\algorithmicindent} {\bfseries end for}
    \STATE\hspace{\algorithmicindent} {\bfseries update} discriminator by  $u$ times. 
    \STATE {\bfseries end for}
   \STATE {\bfseries return}: $G $

\end{algorithmic}
\end{algorithm}

\subsection{Targeted Attack}
Next we extend the proposed algorithms to conduct targeted attacks. Given an adjacency matrix and a feature matrix, the goal is to make nodes to be classified as a desired class by adding fake nodes. 
Assume our goal is to attack a subset of nodes $S$ and $y_i^*$ is the target label for node $i$, the attack objective function can be defined as: 
\begin{equation}
\!\! J(A' , X') \!= \!
    \! \sum_{i\in S} \!
   \left(\! [f(X', A')]_{i, y_i^*}\! - \!
\max\! \Big(\![f(X', A')]_{i,:}\! \Big)\!\right) \!, 
\end{equation}
In this objective function, if the largest logit of node  $i$ is  the target label $y_i^*$, the objective function  value will be  $0$; otherwise the value is negative.
Similar to the non-targeted attacks, we would like to find $B$, $C$ and $X_{fake}$ using Greedy attack to solve the optimization problem (\ref{eq:greedy}) and Greedy-GAN attack to solve the optimization problem (\ref{eq:greedy_gan}).

In our experiments, we consider two cases: attacking a group of randomly sampled nodes ($S=\{v_i, \dots, v_m\}$)
and attacking one node (only one element in $S$). 
For attacking a group of nodes, the fake nodes labels are given in two ways: (1)  uniform distribution, which is the same as in the non-targeted attack setting; (2) using the target labels $y_i^*$.
For attacking a single target node, we add three fake nodes with the target label. 

\section{Experimental Results}
\label{experiment}
We use Cora (2708 nodes, 5429 edges, 1433 features) and Citeseer (3312 nodes, 4732 edges, 3703 features) attribute graphs as benchmarks. In all the experiments, we split the data into 10\% for training, 20\% for validation, and 70\% for testing. 
For non-targeted attacks, the results are evaluated by GCN classification accuracy \MH {move the lower the better into corresponding table title}(the lower the better), while for targeted attacks the results are evaluated by the attack success rates (the higher the better). 
Also, we conduct two kinds of experiments: attacking a single node and attacking a group of nodes. \MH {??? I think we have deleted attacking the whole testing dataset?} \XY{attacking the whole datasets need to add too many edges, it will cause number of cc change largely.} 
\XY{that is using a discriminator to detect feature difference between fake and real nodes. Do you want me to put on table 2 ?}
For attacking a single node, we add three fake nodes; for attacking a group of 100 nodes, we add 2.5\% nodes as the fake nodes. To show unnoticeable perturbations on graph structures we not only use 95\% confidence interval of nodes degrees distribution, but also number of connected components of the graphs. To show our features are like the real ones, we use F-1 score of a classifier.

We include the following algorithms into  comparisons:
\begin{itemize}
\item Greedy and Greedy-GAN: our proposed algorithm that attack GCN by adding fake nodes.
\item GradArgmax ( \cite{pmlr-v80-dai18b}):  white box GCN attack by only changing links of graphs\footnote{The original GradArgmax does not work for adding fake nodes, but we modify them to do so by 1) initialize fake nodes with random features 2) add the constraints to their optimization to restrict them to modify only those links. }. 
\item Nettack( \cite{netattack}): state-of-art data poisoning GCN attack by changing existing nodes' attributes and links.
\MH{list all the methods in the table}\item Random: randomly generating features of fake nodes, and randomly adding links between fake nodes and real ones. We add this method as a baseline.
\end{itemize}


\subsection {Non-targeted Attack} 

We compare the effectiveness of our methods with random and GradArgmax algorithms \cite{pmlr-v80-dai18b}. 

For attacking a single node, three fakes nodes with an average 5 edges are added. 
Table \ref{tab:non-targeted_one} shows that Greedy method achieves lower accuracy than GradArgmax method in both types of normalization. 
In our experiments
the number of added links is small, thus the fake nodes' degrees are all fell in 95\% confidence interval of power-law distribution. To further illustrate that our method perturbs the graph very slightly, we use number of connected components which is a stronger criterion than node degree distributions. \CH{number of strongly connnected components?}After adding fakes nodes and edges, the number of connected components only changes within 3\% \MH {get a number instead fo saying slightly }\XY{changed} for both datasets. 
Also to be noted that we did not perform Greedy-GAN attack for one node, because the numbers of real and fake nodes are so unbalanced and the discriminator will over-fit to the added three nodes.

\begin{table}
    \centering
    \caption {Accuracy for non-targeted attacks on a single node with different normalization in Section \ref{prelim} .
        The last two rows are average number of connected components for each attacking case.}
    \resizebox{0.46\textwidth}{!}{\begin{tabular} {lllll} 
        \hline
        Dataset& \multicolumn{2}{c}{Cora}  &\multicolumn{2}{c}  {Citeseer} \\
        \hline
        Normalization & row-wise  & symmetric  & row-wise  & symmetric \\ 
        \hline
        Greedy & 0.09 & 0.08 & 0.11 & 0.09 \\
        GradArgmax & 0.10 & 0.56 & 0.15 & 0.55 \\
        \hline
        CC of graph & 78 & 78 & 438 & 438 \\
        Avg CC of Greedy & 77.01 & 76.96 & 435.11 & 434.67 \\
        Avg CC of GradArgmax & 77.14 & 76.55 & 435.72 & 428.62 \\
        \hline
    \end{tabular}}

    \label{tab:non-targeted_one}
\end{table}

Next we consider attacking a group of nodes together rather than a single node. 
In the experiment, we randomly sampled 100 nodes as our attack target and added 2.5\% labelled fake nodes with average 10 links per fake node. The results  in Table \ref{tab:attack_100} shows that our Greedy and Greedy-GAN methods perform better than GradArgmax. Moreover, even directly compared with the results from the Nettack paper \cite{netattack}, which changing edges and features on existing nodes simultaneously, our methods could reach comparable result in Table \ref{tab:non-targeted_one} and \ref{tab:attack_100}.\CH{a bit confusing since we also have GradArgmax in the table and we are much better}\XY{that is because the gradArgmax does not give every fake node an edge to the real one, some fake nodes are still dangling nodes, that they are not good, changed} 

For the Greedy-GAN method, we choose the parameter c=0.1 in Equation \ref{eq:greedy_gan}, in the way that it’s more focusing on attacking than generating real features.
We notice that there are some cases that Greedy-GAN performances better than Greedy Algorithm in terms of accuracy of GCN, which is a little bit against people intuition. Intuitively, Greedy should always be better than Greedy GAN, since Greedy-GAN limits the feature spaces. We observe that Greedy method is easier to stuck in local optimum during the optimization procedure (e.g., keep adding/dropping the same edge), while Greedy-GAN has more smoothed optimization curve, because of the added discrimination loss. We will discuss more about how different $c$ influences efficiency of attacking in Section \ref{diff_c_subsect}. 

To check whether the attacks change the graph structure significantly, we check the number of connected components. Number of connected components is a widely used parameter in graph theory when describing graphs structures. It is a stronger criterion than node degree distributions. 

We observed that GradArgmax attack produce a larger number of components, while both of our methods stay relatively unchanged.   
Furthermore, we train a classifier using randomly generated nodes, then feed it with fake features generated by Greedy and Greedy-GAN algorithms. The f1 scores of the classifier indicate that it is much harder to differentiate the features of a node between real and fake under attacks by Greedy-GAN, as compared to Greedy.
This means that Greedy-GAN attack, although sometimes having lower success rate, can produce fake nodes that are harder to be detected.

\begin{table}
    \centering
    \caption {Accuracy of GCN before and after non-targeted attacks for attacking random 100 nodes in each graph. Each graph added 2.5 \% fake nodes. Each fake nodes is with on average 10 fake links. Note that the final two rows are f1 score of a classifier---lower values indicate added fake nodes are harder to be detected.}
    \resizebox{0.46\textwidth}{!} {\begin{tabular} {lllll} 
        \hline
        Dataset& \multicolumn{2}{c}{Cora}  &\multicolumn{2}{c}  {Citeseer} \\
        \hline
        Normalization & row-wise  & symmetric  & row-wise  & symmetric \\ 
        \hline
        Clean & 0.84 & 0.81 & 0.76 & 0.73  \\
        Random & 0.77 &0.79 & 0.68 & 0.69\\
        GradArgmax & 0.53 &0.61 & 0.55  & 0.47\\
        Greedy & 0.08 & 0.03 & 0.03 & 0.04 \\
        Greedy-GAN & 0.05& 0.09 & 0.07 & 0.03 \\
        \hline
        CC of GradArgmax & 74 & 118 & 439 & 457 \\
        CC of Greedy & 71 & 72 & 418 & 401 \\
        CC of Greedy-GAN & 72 & 70 & 419 & 420 \\
        \hline
         F1 score for Greedy & 0.73 & 0.67 & 0.64 & 0.77\\
         F1 score for Greedy-GAN & 0.35 & 0.47  & 0.43 & 0.65 \\
        \hline
    \end{tabular}}

    \label{tab:attack_100}
\end{table}

\subsection{Targeted Attack}

\begin{figure}[!tbp]
    \includegraphics[width=0.46\textwidth]{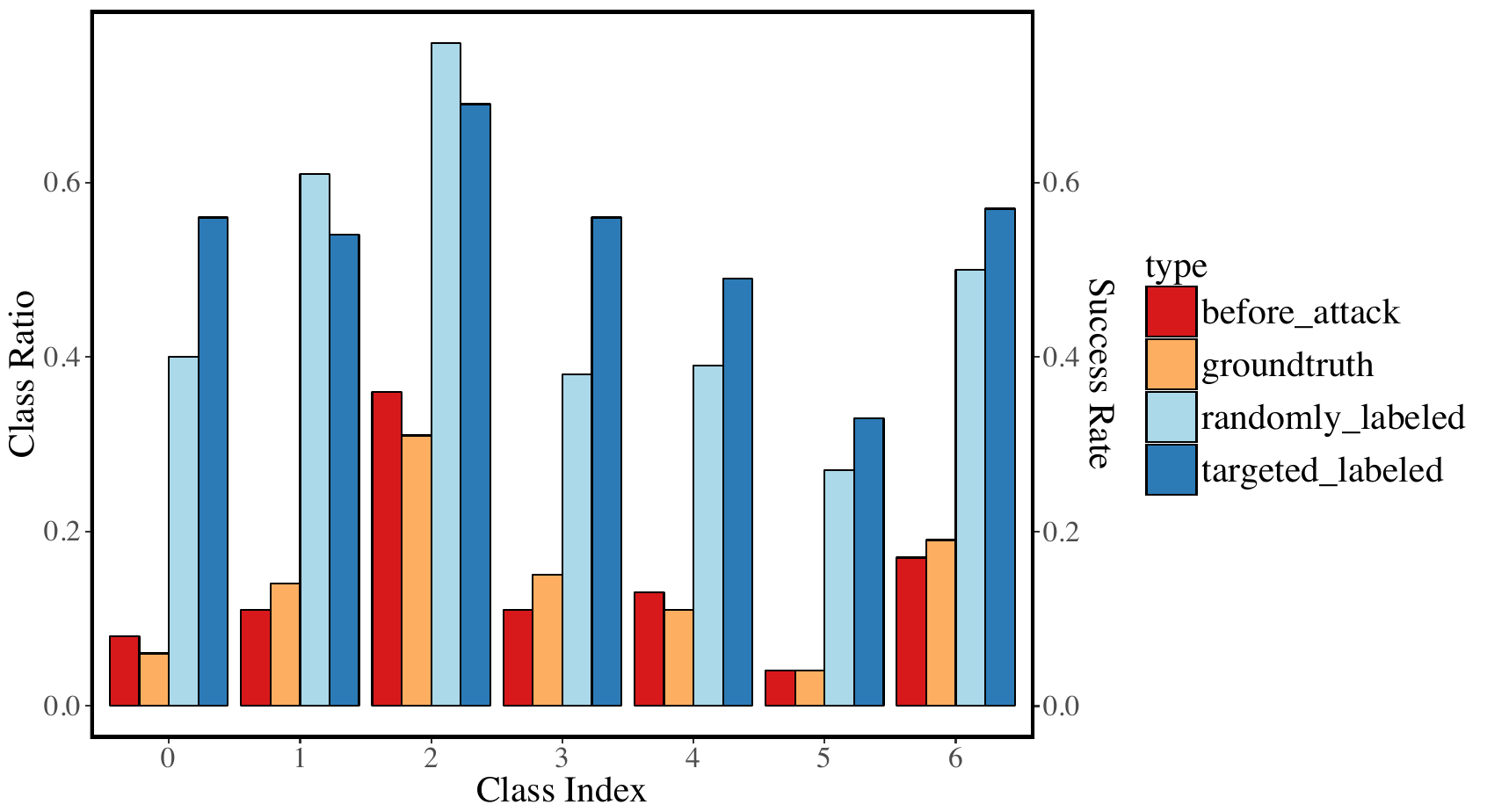}
    \caption{Row normalized Cora dataset. Right axis: Success rate of targeted attacking a group of 100 nodes with randomly and targeted labelling methods;
    Left axis: Groundtruth and before attacking label distributions; 
    with 7 different classes.}
    \label{fig:cora_target_attack}
\end{figure}

\begin{figure}[!tbp]
    \includegraphics[width=0.46\textwidth]{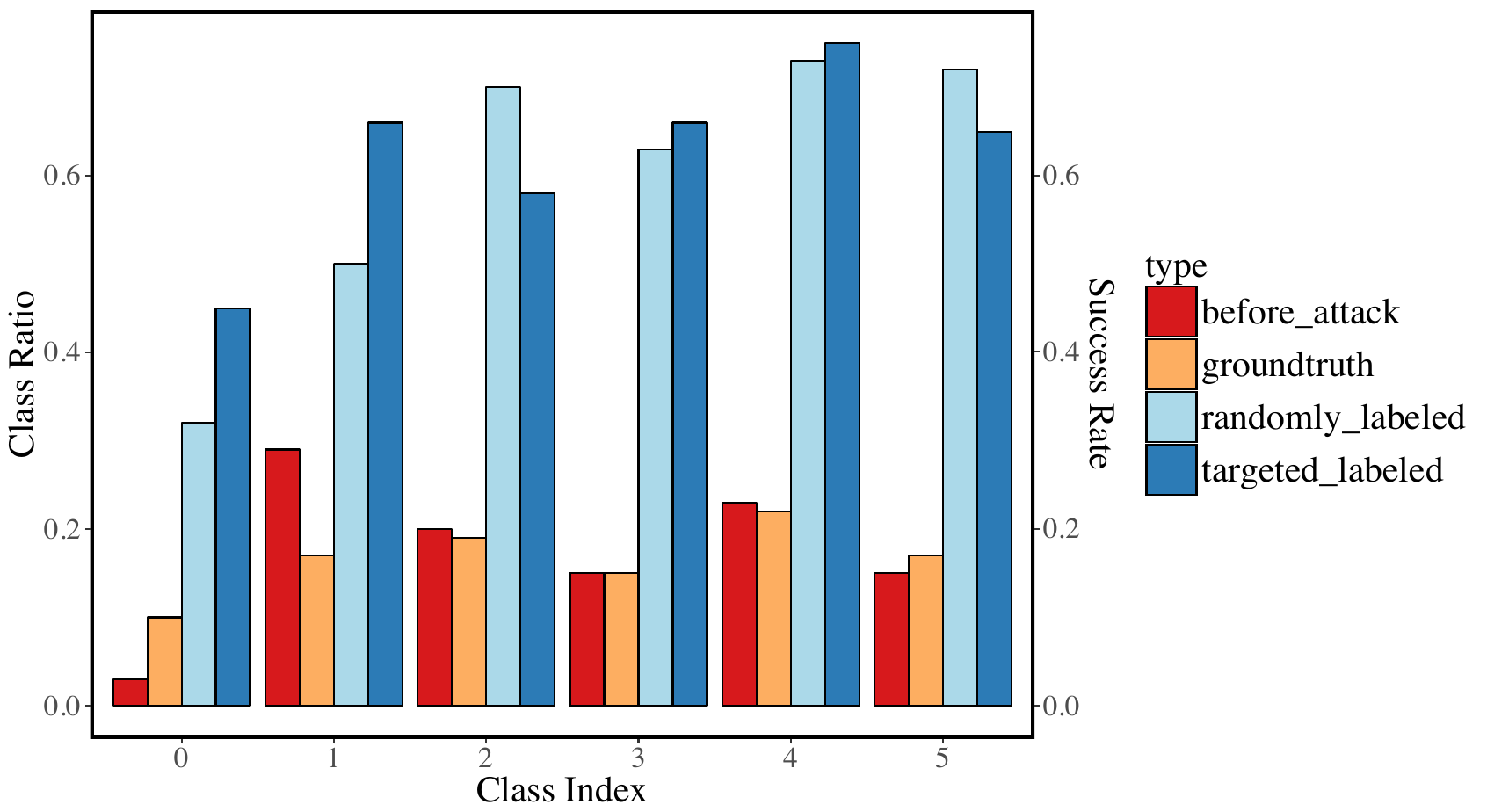}
    \caption{Row normalized Citeseer dataset. Right axis: Success rate of targeted attacking a group of 100 nodes with randomly and targeted labelling methods;
    Left axis: Groundtruth and before attacking label distributions; 
    with 6 different classes. }
    \label{fig:citeseer_target_attack}
\end{figure}

The targeted attack tries to fool the GCN to classify specific nodes to target classes. We perform two different experiments for targeted attacks: (1) attack only a single node; (2) attack a group of 100 nodes on test data.  In our experiments, we use the same setting as non-targeted attack for the number of fake nodes and edges.

\begin{table}
    \centering
    \caption{success rate of targeted attacks on a singe node for different targeted classes  using Greedy Attack. \CH{What is class index (and it should not be put there), also, make sure the first character is capitaled..}}
    \resizebox{0.46\textwidth}{!}{\begin{tabular} {lllllllll} 
        \hline
        & & \multicolumn{6}{c}{Class Index} \\
        \hline
        Dataset& Normalization & 0 & 1 & 2 & 3 & 4 & 5& 6 \\
        \hline
        Cora & row-wise & 0.90& 0.87 & 0.74 & 0.78 & 0.81 & 0.89 & 0.80 \\
        Cora & symmetric& 0.90& 0.86 & 0.75 & 0.80 & 0.87 & 0.86 & 0.81 \\
        Citeseer &row-wise& 0.80 & 0.81 & 0.80 & 0.83 & 0.75 & 0.80 &\\
        Citeseer &symmetric& 0.83 & 0.80 & 0.79& 0.83& 0.73& 0.81 &\\
        \hline
    \end{tabular}}
    
    \label{tab:targeted_one}
    \vspace{-5pt}
\end{table}

When attacking only a single node, we add three fake nodes with target labels. Table \ref{tab:targeted_one} shows that the success rate of our algorithm is around 80\% on different target labels. The addition of three nodes produced no noticeable change in label distributions. 

We also conduct targeted attacks on groups of nodes. 
In Figure \ref{fig:cora_target_attack} and \ref{fig:citeseer_target_attack}, we notice that attacking nodes to certain target classes are more difficult than others. \CH{not clear, also, which table or figure to look at?}For example, in Figure \ref{fig:citeseer_target_attack}, targeted attack has a lower success rate with a target class 0 (249 nodes) from Citeseer dataset (3312 nodes in 6 classes). 
 Success rate of targeted attack is positively correlated with the percentage of that class in the ground truth. 
  The reason might be that if there are less nodes belonging to a class,  the attacked nodes only have weak linkage to that class,  and the classification is more influenced by nodes in other classes. This makes it harder to turn the predicted label to class with few nodes.
For fake nodes labels, we do experiments on both randomly and targeted labels. Targeted labeling generally gives better result. Table \ref{tab:targeted} shows that targeted attacks success rate for Greedy and Greedy-GAN algorithms using targeted labeling. 


\begin{table}
    \centering
    \caption {Success rate for targeted attacks on the 100 nodes on graph.}

    \resizebox{0.46\textwidth}{!}{\begin{tabular} {lllll} 
        \hline
        Dataset& \multicolumn{2}{c}{Cora}  &\multicolumn{2}{c}  {Citeseer} \\
        \hline
        Normalization & row-wise & symmetric & row-wise  &  symmetric  \\ 
        \hline
        Greedy & 0.69 & 0.66 & 0.75 & 0.68 \\
        Greedy-GAN & 0.76 & 0.53 & 0.78 & 0.65 \\
        \hline

    \end{tabular}}

    \label{tab:targeted}
\end{table}

\subsection{ Data Poisoning }
In industry, GCN is normally applied to longitudinal data. Due to the time-consuming process of training a new network, networks are periodically retrained instead. We thus consider the scenario where the GCN is re-trained under the dataset with fake nodes. This is also known as the data poisoning scenario, while the previous discussed case (without retraining) is called test time attack. 

Table \ref{tab:poisoning} presents the accuracy after non-targeted data poisoning. The GCN is retrained with learning rate 0.01, 50 epochs after the data has been modified. The results show that both attacking and poisoning can effectively reduce the accuracy of GCN, and the symmetric normalization is more robust under poisoning attacks overall. 
\begin{table}
    \centering
    \caption {Accuracy after poisoning GCN.}

    \label{tab:method_comparsion }
    \resizebox{0.46\textwidth}{!} {\begin{tabular} {lllll} 
        \hline
        Dataset& \multicolumn{2}{c}{Cora}  &\multicolumn{2}{c}  {Citeseer} \\
        \hline
        Normalization & row-wise  & symmetric  & row-wise  & symmetric \\ 
        \hline
        Nettack 100 nodes &0.01 &0.04 &0.04 &0.07 \\
        Greedy one node & 0.08 & 0.09 & 0.14 & 0.20 \\
        Greedy 100 nodes & 0.06 & 0.04 & 0.01 & 0.08 \\
        Greedy-GAN 100 nodes & 0.01& 0.11 & 0.03 & 0.03 \\
        \hline
    \end{tabular}}
    \label{tab:poisoning}
\end{table}

For specific node data poisoning\CH{poisoning a single node?}\XY{changed}\MH {data poisoning for specific node? if that is the case, the former should be changed data poisoning for all nodes}\XY{data poisoning for a specific group of nodes. For Zugner's paper did he poison all graph? I did not poisoning the whole graph, I only measures the target nodes accuracy}, since only three fake nodes are added, we did not change existing edges in the training dataset, thus the 'good' nodes and edges in the training dataset are still dominating. While for a group of nodes, it works quite well, we could reach as low as 0.01 accuracy which is as good as the original Nettack \cite{netattack} by changing features and edges directly and retraining the network until converge, and our result on Citeseer is even better. 

\subsection{Degree of Nodes} 

\begin{table}
    \centering
    \caption {An typical case for nodes with different degrees, under Greedy non-targeted attack and poisoning; Citeseer Dataset with row normalization, seed = 42.}
    \resizebox{0.46\textwidth}{!}{\begin{tabular}{llllll} 
        \hline
        nodes degree & (0,5] & (5,10] & (10,20] & (20, $\infty$) & total accuracy \\
        \hline
        clean & 0.78 & 0.83 & 1.0 & 1.0 & 0.80 \\
        Greedy Attacking &0.037& 0.25 & 0.50 & 1.0 & 0.10\\
        Greedy Poisoning & 0.025 & 0.417 & 0.667 & 1.0 & 0.12 \\
        Greedy-GAN Attacking & 0.037 & 0.25 & 0.67 & 1.0 & 0.11 \\
        Greedy-GAN Poisoning  & 0.074& 0.33& 0.67& 1.0 & 0.15 \\
        \hline
    \end{tabular}}
       
    \label{tab:degree}
\end{table}
Previous work \cite{netattack} found that nodes with smaller degrees are more vulnerable in data poisoning. Our experiments confirm that attacking a group of nodes and single node also follow that rule. 
Table \ref{tab:degree} presents typical nodes accuracy by degree.
When adding extra links between fake nodes and real nodes, higher degree nodes are more resistant to the impact of fake nodes. And this works for both attacking and poisoning. 


\subsection{Row-wise vs. Symmetric Normalization. }
Although row-wise or symmetric normalization could achieve similar performance in classification, we found their robustnesses vary.
For row-wise normalization, when adding a new edge $e_{i,j}$ on graph, the two elements in adjacency matrix $A_{i,j}$ and $A_{j,i}$ changes from 0 to 1. Therefore, we could get the corresponding two elements in the normalized matrix $\hat A$ changing from 0 to $\hat A_{i,j}^{(t+1)} = 1/(d_{ii} +1)$ and $A_{j,i}^{(t+1)} = 1/(d_{jj} +1)$. Also, other nonzero elements $\hat A_{i, .}^{(t)}, \hat A_{j, .}^{(t)} $ changes to
\begin{align*}
\hat A_{i,.}^{(t+1)} ={ d_{ii}^{(t)} }/({d_{ii}^{(t)}+1})  \hat A_{i, .}^{(t)} \\
\hat A_{j,.}^{(t+1)} ={ d_{jj}^{(t)} }/({d_{jj}^{(t)}+1})  \hat A_{j, .}^{(t)}
\end{align*}

For symmetric normalization, origin elements $\hat A_{i,j}^{(t)}$ and $\hat A_{j,i}^{(t)}$ in the original normalized matrix changes from 0 to 1/ $\sqrt{(d_{ii}^{(t)}+1)(d_{jj}^{(t)}+1)}$ while other nonzero elements $\hat A_{i, .}^{(t)}$, $\hat A_{j, .}^{(t)} $ changes to
\begin{align*}
\hat A_{i,.}^{(t+1)} = 
\hat A_{.,i}^{(t+1)} ={ \sqrt{ d_{ii}^{(t)} }}/{\sqrt{ d_{ii}^{(t)}+1}}  \hat A_{i, .}^{(t)} \\
\hat A_{j,.}^{(t+1)} =
\hat A_{.,j}^{(t+1)} ={ \sqrt{ d_{jj}^{(t)} }}/{\sqrt {d_{jj}^{(t)}+1}}  \hat A_{j, .}^{(t)} \\
\end{align*}
To sum up, on one hand, the magnitude of change is greater in row normalized elements than in symmetric normalized elements.
On the other hand, when adding or dropping an edge in the graph, only 2 rows are affected in the row normalized adjacency matrix. However for symmetric case, it affects both 2 rows and 2 columns. 

Therefore, we find that when adding smaller amount of edges, number of affected rows/columns is dominated so that symmetric normalized GCN is more vulnerable. When adding larger amount of edges, the magnitude of change in elements plays a more important role so that the row normalized GCN is more vulnerable. For example, in Table \ref{tab:non-targeted_one}, the average edges per fake node is 5, and accuracy of symmetric normalized GCN is smaller than row normalized one. While in Table \ref{tab:compare} symmetric normalized GCN is more robust in all of the methods, where we add 5 \% fake nodes to attack the whole test dataset of graph (70 \% nodes of graph.) by adding 10,000 fake edges (around 60 to 75 edges per fake node).  For attacking a group of 100 nodes, the average number edges per fake nodes is 10, which causes those two effects contribute equally. Therefore, there is no significant differences in terms of robustness.

\begin{table}
    \centering
    \caption{ An extreme case of adding 5 \% fake nodes and 10,000 edges for non-targeted
    attacking the whole test dataset (70 \% nodes).  }
    \resizebox{0.46\textwidth}{!}{\begin{tabular} {lllll} 
        \hline
        Dataset& \multicolumn{2}{c}{Cora}  &\multicolumn{2}{c}  {Citeseer} \\
        \hline
        Normalization & row-wise  & symmetric  & row-wise  & symmetric \\ 
        \hline
        Clean & 0.84 & 0.81 & 0.76 & 0.75  \\
        Random & 0.34 &0.42 & 0.49 & 0.51\\
        Greedy & 0.11 & 0.15 & 0.02 & 0.16 \\
        Greedy-GAN & 0.11 & 0.25 & 0.07 & 0.17 \\
        \hline
        
    \end{tabular}}
    
    \label{tab:compare}
\end{table}


\begin{figure}[!tbp]
    \includegraphics[width=0.46\textwidth]{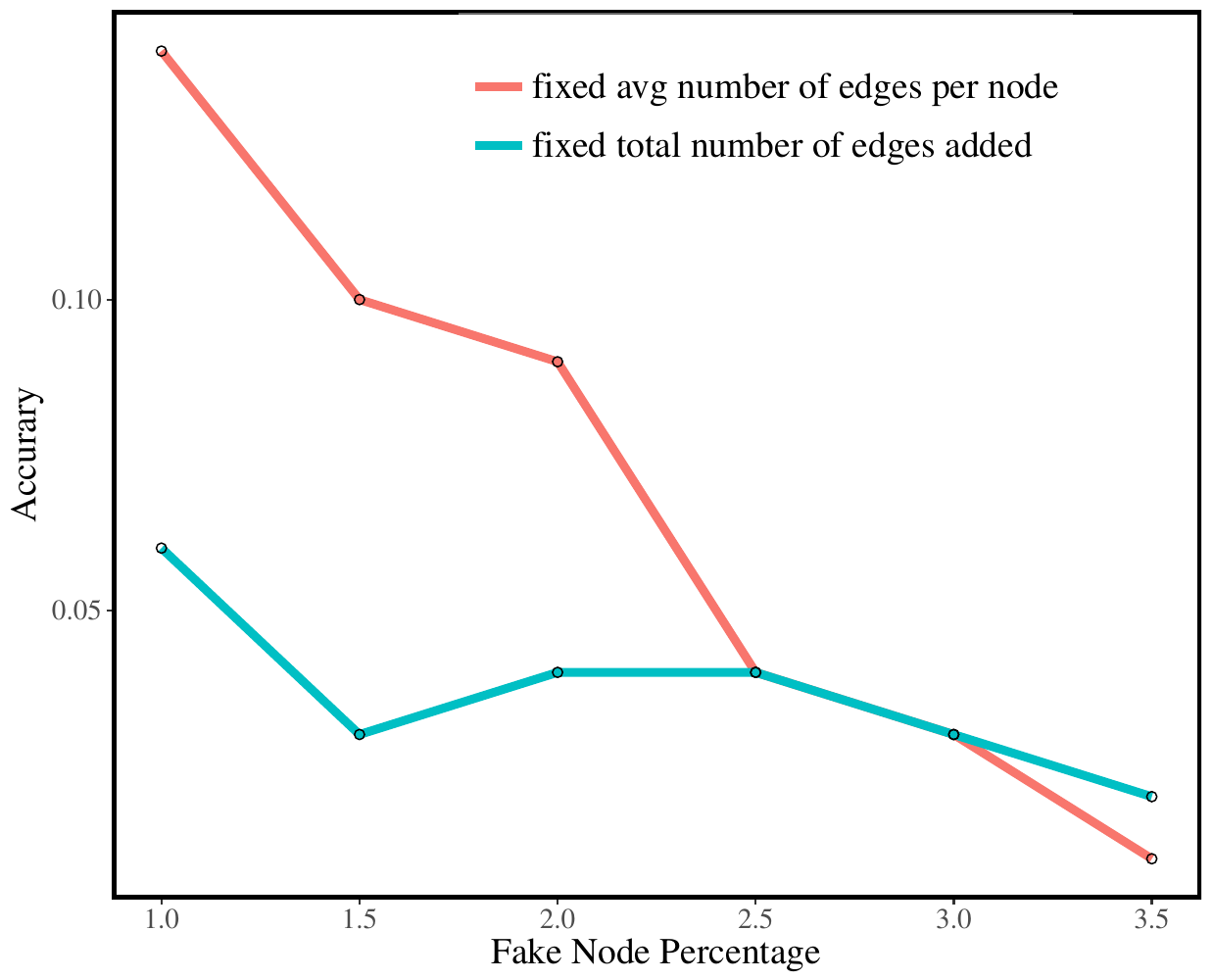}
    \caption{Accuracy varies with different number of fake nodes and different number of added edges, using symmetric Citeseer dataset ; Red line: Fixed average degree of fake nodes as 10; Blue line: Fixed total number of added edges. }
    \label{fig:diff_fake_nodes}
\end{figure}

\subsection{Number of Fake Nodes. }
In this part, we explore how the number of fake nodes influences the efficiency of attacking. We attack a group of 100 nodes with different number of fake nodes. Figure \ref{fig:diff_fake_nodes} shows how the number of fake nodes influences the classification accuracy. We use use 1\%,1.5\%,2 \%, 2.5\% and 3\% of fake nodes, and assign random labels to the fake nodes. 
As expected, more fake nodes yields more effective attacks. We notice that the number of added edges between fake nodes and real nodes also plays a important role. The blue line in Figure \ref{fig:diff_fake_nodes} ranges from 2\% to 6\% with adding same total amount edges (as same as the experiment in  Section 5.1 and 5.2), with different number of fake nodes. This indicates even with very small amount of fake nodes, we can attack the accuracy of GCN to very low, for example, if we only 1 \% fake nodes, with average degree of fake node 25, in Figure \ref{fig:diff_fake_nodes}, the accuracy is down to 0.06. 
On the other hand, if we fixed the average degree of fake nodes as 10 (also the same as Section 5.1 and 5.2 average degree of fake nodes setting), the accuracy varies from 0.14 to 0.01 with the number of fakes nodes increasing from 1 \% to 3\%. 

\subsection{Different $\textbf{c}$ in Greedy-GAN}
\label{diff_c_subsect}
\begin{figure}[!tbp]
    \includegraphics[width=0.46\textwidth]{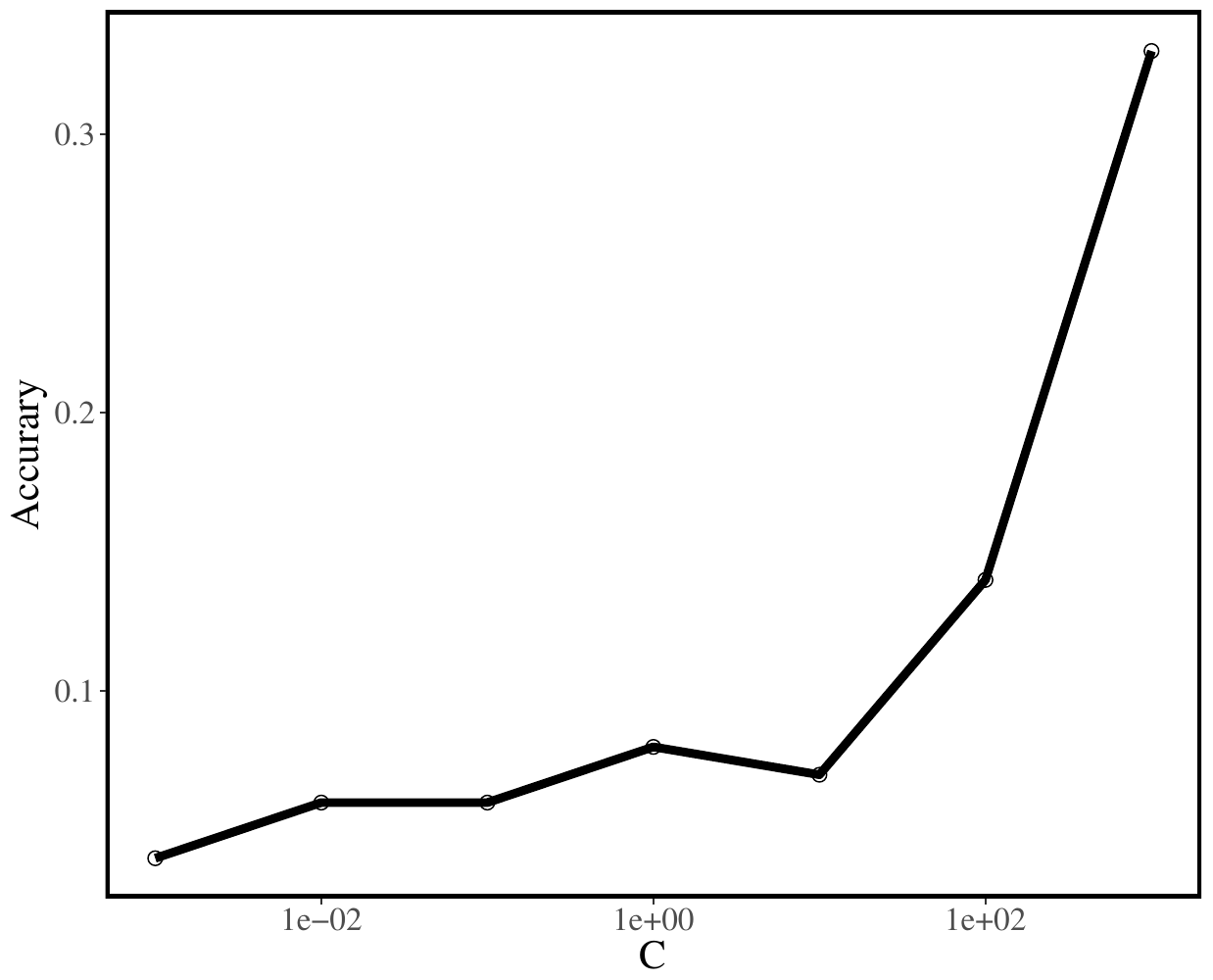}
    \caption{Row normalized Cora dataset: Accuracy with difficult $c$ in Algorithm \ref{alg_cw}}
    \label{fig:diff_c}
\end{figure}

We study how different c in Algorithm \ref{alg_cw} influences the attack efficiency. In Section 4.1, we claimed that 
if $c$ is large, the objective function is dominated by the discriminator, so the node features generated will be very close to real ones but with a lower attack successful rate (higher accuracy). If $c$ is greater than 1, we say realness of the nodes are more important, if $c$ is smaller than 1, then the system is more on attacking.
In Figure \ref{fig:diff_c}, $c$ varies from very small value 0.001 to very large value 1000 increasing by 10 times each time, there is a positive correlation between $c$ and accuracy of GCN. We notice that when $c$ is smaller or equal to 10, the accuracy is varies from 0.04 to 0.08, which means even the systems is more on realness of the fake nodes features ($c>1$), the we could get efficient attacking using Greedy-GAN. 
Greedy-GAN algorithm could maintain realness of fake nodes without losing attacking efficiency.  Unless $c$ is extremely large, say 1000, the accuracy 0.33, which is not as efficient as the result in Table \ref{tab:attack_100} in terms of attacking.

\subsection{Attacking Large Scale Graphs}
When training large scale graphs, original GCN will not work due to memory bottleneck. Thus algorithms based on neighbourhood sampling come out the relief this issue \cite{fastgcn} \cite{graphsage}. Because there is a neighbourhood sampling function in these algorithms, directly adding or deleting edges on the original graphs method become less effective than on GCN. The reason is when training GraphSAGE (or other large scale graph neural networks), sampling neighbourhood could be view as dropping edges during training, \cite{pmlr-v80-dai18b} shows that use dropping edges while training  is a cheap method to increase the robustness of GCN. As the result, attacking GraphSAGE (or other large scale graph neural networks) is more difficult than attacking GCN.  

In this section, we use a GCN aggregator for GraphSAGE for nodes prediction. We use Reddits dataset (232,965 nodes, 11,606,919 edges,41 classes and 602 features) as a benchmark. For labeling rate, we use the same setting as GraphSAGE\cite{graphsage} paper. When considering the attacking single nodes scenario, for changing edges directly on the existing graphs, the accuracy of GraphSAGE drop from 0.98 to 0.95.
For adding fake nodes the accuracy drops to 0.07. Reddits dataset has continuous feature space, thus it would be more vulnerable for fake nodes attacks than the discrete feature space datasets (Cora and Citeseer). Adding fake nodes are easily to introduce new features to the target-attacked node in continuous features space, thus it will lead to accuracy dropping. 

\section{Conclusion}
\label{conclusion}
In this paper, we develop two algorithms, Greedy and Greedy-GAN, to attack GCNs by adding fake nodes and without changing any existing edge or feature. 
Our experimental results show that both algorithms can successfully attack GCNs. To make the attack unnoticeable, we added a discriminator using the Greedy-GAN algorithm to generate features of fake nodes. Furthermore, we explored parameter sensitivities on degree of nodes; different normalize methods to evaluate the effectiveness of our attacks; different number of fake nodes and the trade-off between realness of features and attacking efficiency. Moreover we scaled up our attacks to large scale graph data that uses GraphSAGE\cite{graphsage}. 

In future, we would like to explore the defence of adversarial attacks on GCNs as well as on large scale networks such as GraphSAGE\cite{graphsage}.



\bibliographystyle{ACM-Reference-Format}
\bibliography{citation2}

\end{document}